\documentclass[conference]{IEEEtran}
\IEEEoverridecommandlockouts
\usepackage{cite}
\usepackage{amsmath,amssymb,amsfonts}
\usepackage{algorithmic}
\usepackage{graphicx}
\usepackage{textcomp}
\usepackage{xcolor}
\usepackage{cleveref}

\newcommand{\introduce}[2]{\textit{#1} \textbf{(#2)}}
\newcommand{\highlight}[2]{\textit{#1} (\textit{#2})}
 \newcommand{\linebreakand}{%
      \end{@IEEEauthorhalign}
      \hfill\mbox{}\par
      \mbox{}\hfill\begin{@IEEEauthorhalign}
}

\def\BibTeX{{\rm B\kern-.05em{\sc i\kern-.025em b}\kern-.08em
    T\kern-.1667em\lower.7ex\hbox{E}\kern-.125emX}}
\begin{document}

\title{Landscape of AI safety concerns - A methodology to support safety assurance for AI-based autonomous systems\\
}

\author{
\IEEEauthorblockN{Ronald Schnitzer}
\IEEEauthorblockA{\textit{Technical University of Munich} \\
\textit{Siemens AG}\\
Munich, Germany \\
ronald.schnitzer@tum.de}
*Corresponding author
~\\
\and
\IEEEauthorblockN{Lennart Kilian}
\IEEEauthorblockA{\textit{Siemens AG} \\
Munich, Germany \\
lennart.kilian@siemens.com}

~\\
\and
\IEEEauthorblockN{Simon Roessner}
\IEEEauthorblockA{\textit{Siemens AG} \\
Munich, Gremany \\
simon.roessner@siemens.com}

~\\
\and
\IEEEauthorblockN{Konstantinos Theodorou}
\IEEEauthorblockA{\textit{Fraunhofer IKS} \\
Munich, Germany \\
konstantinos.theodorou@iks.fraunhofer.de}

~\\
\and
\IEEEauthorblockN{Sonja Zillner}
\IEEEauthorblockA{\textit{Technical University of Munich} \\
\textit{Siemens AG}\\
Munich, Germany \\
sonja.zillner@tum.de}}

\maketitle

\begin{abstract}
Artificial Intelligence (AI) has emerged as a key technology, driving advancements across a range of applications.
Its integration into modern autonomous systems requires assuring safety. 
However, the challenge of assuring safety in systems that incorporate AI components is substantial.
The lack of concrete specifications, and also the complexity of both the operational environment and the system itself, leads to various aspects of uncertain behavior and complicates the derivation of convincing evidence for system safety.
Nonetheless, scholars proposed to thoroughly analyze and mitigate AI-specific insufficiencies, so-called AI safety concerns, which yields essential evidence supporting a convincing assurance case.
In this paper, we build upon this idea and propose the so-called \textit{Landscape of AI Safety Concerns}, a novel methodology designed to support the creation of safety assurance cases for AI-based systems by systematically demonstrating the absence of AI safety concerns.
The methodology's application is illustrated through a case study involving a driverless regional train, demonstrating its practicality and effectiveness.
\end{abstract}


\section{Introduction}
\label{sec: Intro}
Recent advancements in \highlight{Artificial Intelligence}{AI}, particularly in \highlight{Machine Learning}{ML}, have significantly enhanced a multitude of applications.
The ability of AI to solve complex tasks in challenging environments and even exceed human performance in some tasks has been a deciding factor for its integration into autonomous systems. 
However, due to the potential harm caused by such systems in safety-critical applications, assuring their safety is fundamental.
Safety assurance can be demonstrated through the development of a safety assurance case, which, according to ISO/IEC/IEEE 15026-1, is a "reasoned, auditable artefact created to support the contention that its top-level claim (or set of claims) is satisfied, including systematic argumentation and its underlying evidence and explicit assumptions that support the claim(s)" \cite{ISO15026-1}. 
However, the integration of AI/ML-based \footnote[1]{Since ML, being a subset of AI, is the dominant technology used in safety-critical applications, but the methodology outlined in this work is not limited to ML in particular, we will use the terms AI and ML interchangeably in the following.} components in autonomous systems introduce additional complexities in developing an assurance case, as opposed to conventional software.

A fundamental challenge in assuring the safety of AI-based systems is caused by the \textit{semantic gap}, which refers to the discrepancy between the intended functionality and the specified functionality of a system \cite{burton_mind_2020}. 
The semantic gap in AI-based systems mainly arises from two factors. 
Firstly, AI-based systems, particularly those based on advanced algorithms such as \highlight{Deep Neural Networks}{DNNs},
adapt a large number of parameters to vast amounts of training data.
These parameters ultimately determine the model's behavior and, consequently, the overall system's performance in operation.
Since developers do not directly specify these parameters and such models are often too complex for their decision-making processes to be comprehended by humans, it is impossible to guarantee that the actual behavior aligns with the intended behavior (e.g., operating safely) in every operational situation in which the safety function is demanded.
Secondly, the complexity of the operational domain contributes to the semantic gap \cite{burton_mind_2020}.
Adequately specifying the operational domain is challenging because the required level of detail is still subject to research.
A conservative approach requires an impractically high number of test samples to achieve a statistically significant estimate of the system's behavior. 
Consequently, applying established testing procedures, such as black-box or requirement-based testing, is rendered impossible. 

Therefore, alternative approaches are needed to demonstrate that AI-based systems' actual behavior aligns with the intended behavior, such as operating safely under all operating conditions.
Researchers have suggested analyzing AI-specific insufficiencies, also referred to as \introduce{AI Safety Concerns}{AI-SCs}, as potential evidence for use in assurance cases \cite{burton_assurance_2022}. 
AI-SCs refer to any AI-related issues that may negatively affect the safety of an AI-based system.
Examples of AI-SCs include \textit{Lack of robustness}, which refers to the insufficient performance of AI-based systems under adverse conditions, and \textit{Problems with synthetic data}, highlighting the risk that AI-based systems trained on synthetic data may not perform as expected in operational settings, potentially leading to hazardous situations.

Although AI-SCs and their related mitigation measures are well discussed in the literature, managing all AI-SCs and producing evidence of their control still presents a significant challenge and requires further research.
This is because AI-SCs manifest at various stages in the lifecycle of an AI-based system and necessitate the application of diverse methods to evaluate and mitigate their impact on system safety.
Furthermore, to guarantee that the evidence produced by these methods is adequate, it is essential to establish a procedure that enables clear evaluation.
This process must determine whether the evidence is sufficiently conclusive.

In summary, a systematic approach is required to collect convincing evidence for the mitigation of AI-SCs.
Expanding upon this idea, we introduce a methodology named \introduce{Landscape of AI Safety Concerns}{LAISC}. 
This methodology is designed to provide comprehensive guidance and consolidate crucial information to systematically develop safety assurance cases for AI-based systems.

The contributions of the paper are:

\begin{itemize}
    \item A methodology to systematically derive evidence for assuring the safety of AI-based systems by demonstrating the absence of AI safety concerns (AI-SCs)
    \item Guidance on how to break down AI-SCs and derive application-specific Verifiable Requirements (VRs)
  
\end{itemize}
To guarantee the method's adaptability, we follow approaches from classical functional safety and demonstrate its feasibility with a case study of a driverless regional train.  

The paper is organized as follows: In \Cref{sec: Related Work}, we outline similar research efforts in the field of AI safety assurance and position this work within the field.
Following this, \Cref{sec: Landscape} describes the LAISC methodology, elaborating on its contribution to the development of safety assurance cases for AI-based systems.
Then, \Cref{sec: Case study} illustrates the application of the methodology to the case study of a driverless regional train, 
\Cref{sec: Discussion} discusses the advantages and limitations of our approach, and \Cref{sec: Conclusion} concludes the paper, offering insights into future research directions.

\section{Related Work}
\label{sec: Related Work}

The related literature for this work is twofold. First, we present an overview of related norms and standards, followed by a discussion on the current state of research in AI safety assurance, positioning our work within this domain.

Safety assurance cases are a widely recognized methodology for transparently demonstrating efforts to ensure the safety of the system under consideration for their intended applications \cite{ashmore_assuring_2022}. Hence, they provide auditors and external parties with a comprehensive overview of safety measures and the system's safety status.
General guidance on assurance cases is provided by the ISO/IEC/IEEE 15026 series \cite{ISO15026}, while IEC 61508 \cite{IEC_61508} emphasizes their significance for electronic safety-related systems.
Domain-specific requirements for assurance cases are often delineated in vertical standards. 
For example, the ISO 26262 series \cite{iso_26262}, which is extended by ISO PAS 21448 \cite{SOTIF}, highlights the necessity of assurance cases in demonstrating compliance with safety requirements in the automotive sector.
Similarly, ISO EN 50129 \cite{EN50129} specifies guidelines for developing safety cases in the railway domain.
Artificial Intelligence is a horizontal technology applicable across various domains.
Employing AI technologies introduces new sources for systematic failures and, thus, unique challenges in system assurance.
Consequently, assuring AI-based systems necessitates methodologies distinct from those in sector-specific standards. ISO/IEC JTC 1 SC 42  and CEN-CENELEC JTC 21 are currently working on several standards in the field of AI.
This includes a document providing guidelines on connecting concepts of functional safety and AI development \cite{iso5469}.

The assurance of AI-based systems is an active field of research, with various aspects being explored \cite{adler_assurance_2022,burton_mind_2020,burton_assurance_2022,hawkins_guidance_2021}.
A thorough survey on assuring ML-based systems is provided by Ashmore et al. \cite{ashmore_assuring_2022}. 
In their work, the authors segment the ML life cycle into four phases, suggesting desiderata for each and discussing available assurance methods and associated challenges. 
The AMLAS process~\cite{hawkins_guidance_2021} also employs the AI life cycle as a framework, defining assurance patterns using \highlight{Goal Structuring Notation}{GSN} \cite{GSN} to derive from top-level safety goals the evidence that needs to be generated in the AI life cycle. 
In this derivation, the AMLAS approach remains generic, as it does not presume specific  AI capabilities and shortcomings. 
In contrast, LAISC supports the development with the progress made in \cite{houben_inspect_2022,schnitzer_ai_2024,willers_safety_2020} on the specific AI properties. 
However, the focus of these works is on the identification of AI-SCs and required actions to manage them, but not on the derivation of a convincing assurance case, which is what we provide in this paper.
Systematic analysis of AI-SCs forms a crucial category of evidence in assurance cases for AI-based systems \cite{burton_assurance_2022}. 
Furthermore, explicitly incorporating AI-SCs into assurance cases provides concrete structuring guidance. 
Utilizing a comprehensive list of AI-SCs, such as the one presented in \cite{schnitzer_ai_2024} (AI-SCs are called \textit{AI hazards} in this work), aids this process.
This approach aligns with current standardization efforts in AI risk management by CEN/CLC JTC 21, which is developing Check Lists for AI Risk Management (CLAIRM).
\section{Landscape of AI Safety Concerns}
\label{sec: Landscape}
The primary goal of a safety assurance case is to demonstrate that the evaluated system achieves an adequate level of safety. 
This demonstration typically involves addressing both: systematic issues and those issues modelled by probabilistic fault models.
Given the unique challenges introduced by AI-based components, particularly systematic issues, LAISC clearly focuses on these.
Based on this consideration, in this section, we propose LAISC, a novel methodology to systematically derive evidence for assuring the safety of an AI-based system by demonstrating the absence of AI safety concerns.

\subsection{LAISC Concept}
\label{subsec: Landscape}

To explain the methodology, we first detail its main components, also depicted in \Cref{fig: Landscape components}.
LAISC, when applied to a specific use case, comprises four key elements:
\begin{figure*}[h!]
    \centering
    \includegraphics[width=0.5\linewidth]{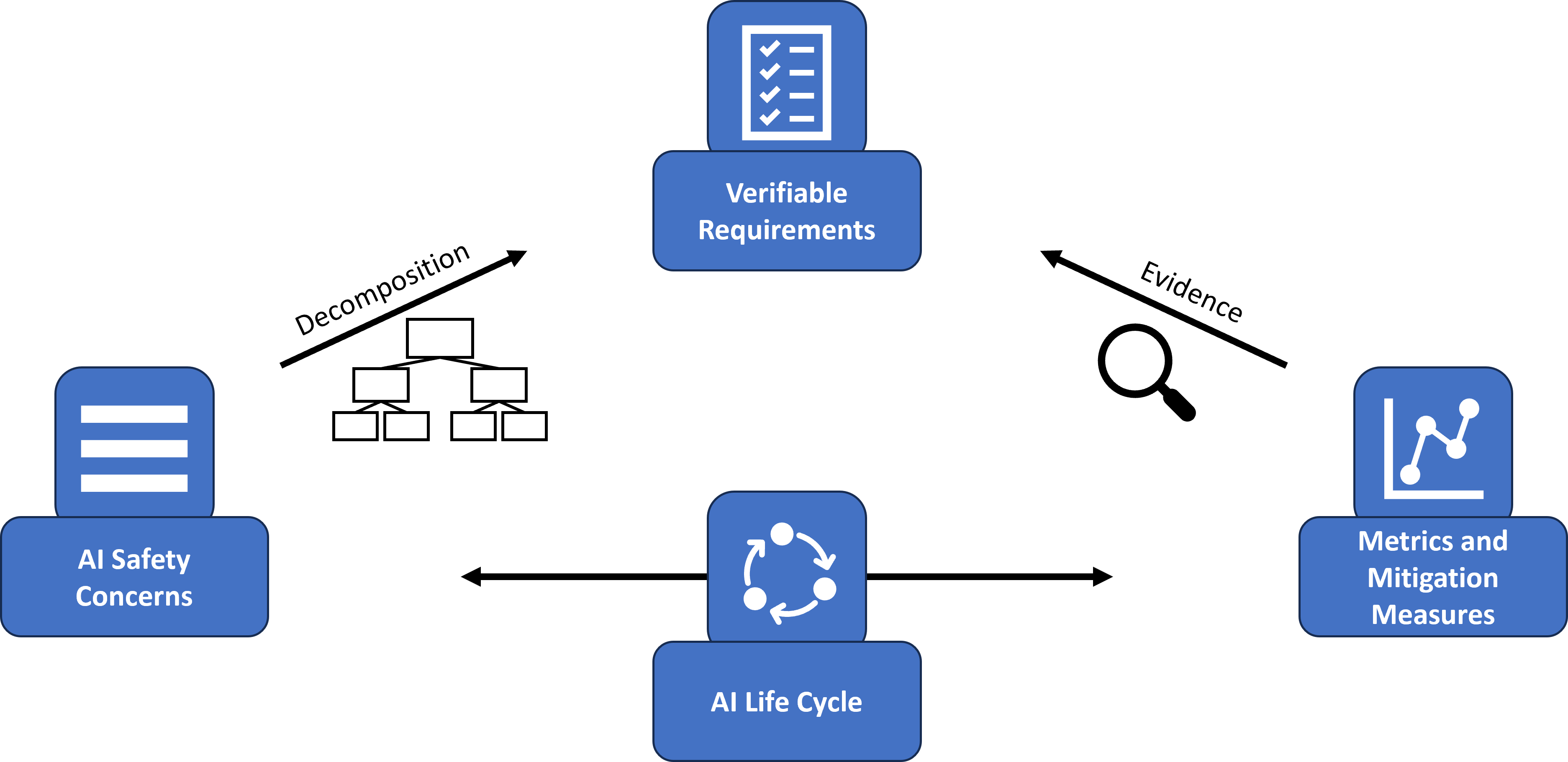}
    \caption{The concept of Landscape of AI Safety 
    Concerns (LAISC).}
    \label{fig: Landscape components}
\end{figure*}

\begin{itemize}
    \item A comprehensive list of AI Safety Concerns (AI-SCs) relevant to the specific use case.
    \item A set of Metrics and Mitigation Measures (M\&Ms) for quantifying and mitigating the impact of AI-SCs and providing evidence for their absence.
    \item The AI life cycle, providing guidance on when to create that evidence during the development and potentially the operation of an AI-based system.
    \item Verifiable Requirements (VRs) that form a structured way to clearly evaluate if the provided evidence is rendered sufficient to show the absence of AI-SCs.

\end{itemize}
\textbf{AI safety concerns:}
In line with \cite{willers_safety_2020}, AI-SCs are defined as \textit{"AI-specific, underlying issues that may negatively impact the safety of a system"}.
Examples of AI-SCs include technical issues such as \textit{Lack of robustness}, which refers to the unreliable behavior of AI models when confronted with changes in operating conditions, and \textit{Lack of explainability}, which pertains to the complex inner workings of such systems that cannot be comprehended by humans. 
Additionally, there are issues like \textit{Poor model design choice} that relate to suboptimal AI-related design decisions, such as the choice of hyperparameters or model class, which can also negatively affect safety.

As illustrated in \Cref{sec: Intro}, assuring the safety of AI systems presents unique challenges compared to conventional software systems. 
AI-SCs represent the gaps introduced by AI technologies.
Therefore, by systematically addressing all AI-SCs, we advocate for bridging this gap and enhancing the safety assurance of AI-based systems. 
Consequently, a comprehensive list of AI-SCs lays the foundation for the LAISC methodology.
\\
\textbf{Metrics and Mitigation Measures:}
The purpose of M\&Ms is to quantify the impact of AI-SCs and providing methods for their mitigation. 
Ultimately, these M\&Ms are envisioned to create the evidence required for the assurance case.
In a specific application, starting with a set of M\&Ms, they are allocated to the relevant AI-SCs.
Having an overview of which AI-SCs are covered by respective M\&Ms helps identify blind spots in terms of AI-SCs that are not yet addressed.
Note that the appropriateness of M\&Ms depends on the particular use case and should be determined by an interdisciplinary team comprising safety, AI, and domain experts.
This is because the requirements for particular M\&Ms  should always be set in relation to the safety requirements of the overall system \cite{picardi_assurance_2020}.
For example, consider assessing the safety-critical performance of a binary semantic segmentation model, which is the AI-based component to be analyzed in \Cref{sec: Case study}. 
This model is designed for identifying a specific object class, such as railroads, within an image by generating a binary segmentation mask.
Here, each pixel is assigned a value of 1 if it represents the targeted object and 0 otherwise.
Implementing performance metrics such as precision or recall is not appropriate since they need to be calculated separately for each pixel.
Instead, a more meaningful metric for semantic segmentation is, for instance, the Intersection over Union (IoU) score \cite{garcia-garcia_survey_2018}.
\\
\textbf{AI life cycle:}
The AI life cycle is crucial to our approach as it determines when   AI-SCs manifest and also the optimal timing for applying M\&Ms to assess and mitigate their impact on system safety.
For systematic mitigation of AI-SCs and generation of evidence for the assurance case, an AI life cycle model is required, which decomposes the entire lifespan of an AI-based system into sub-phases, including the development and operation of its components. 
Various models of the AI life cycle are outlined in the literature \cite{ashmore_assuring_2022,iso_5338_lifecycle,schnitzer_ai_2024,studer_towards_2021,zeller_toward_2024}, differing primarily in detail and granularity.
Which AI life cycle model to use can be decided individually for every use case.
AI-SCs and related activities are distributed along the AI life cycle.
For example, certain AI-SCs pertain to the quality of training data, such as \textit{Inaccurate data labels}, while others arise in post-deployment phases, like \textit{Data drift} \cite{schnitzer_ai_2024}.
This highlights the importance of addressing AI-SCs as early as possible in the life cycle to prevent issue propagation to later stages.
Mitigating AI-SCs as soon as possible in the AI life cycle might additionally reduce the costs for developing AI-based systems and, hence, is a benefit outside the scope of safety.
\\
\textbf{Verifiable Requirements:}
VRs play a crucial role in the LAISC methodology by bridging the gap between abstract AI-SCs and the evidence provided by M\&Ms.
VRs serve as the evaluative criteria to ascertain whether the evidence generated is sufficient to demonstrate the absence of AI-SCs.
This is ensured by enabling a binary (true/false) evaluation based on the evidence produced by the application of M\&Ms. 
Typically, this involves setting specific thresholds when evidence is derived from AI-related metrics based on numerical quantification.
However, relying solely on metric thresholds can be limiting, as not all aspects of AI-SCs are quantifiable numerically.

For AI-SCs that involve more subjective elements, such as \textit{Poor model design choices}, where developers might make suboptimal decisions regarding model attributes like hyperparameters or model class, qualitative assessments become crucial.
In these cases, VRs should incorporate evaluations of the rationale behind such decisions.

For example, a VR for \textit{Poor model design choices} could be formulated as follows: "The model design choices have been justified, and the rationale has been reviewed and approved by at least $n$ independent experts." 
The specific number $n$ should be determined based on the criticality of the application and the potential impact of the design choices on system safety.

Such qualitative VRs might consist of the provision of expert judgment and adherence to well-defined design processes. 
The evidence might include documented justifications of design decisions, peer reviews, or compliance with industry standards.

\subsection{LAISC Process}
\label{subsec: Process}
\begin{figure*}[h!]
    \centering
    \includegraphics[width=0.68\linewidth]{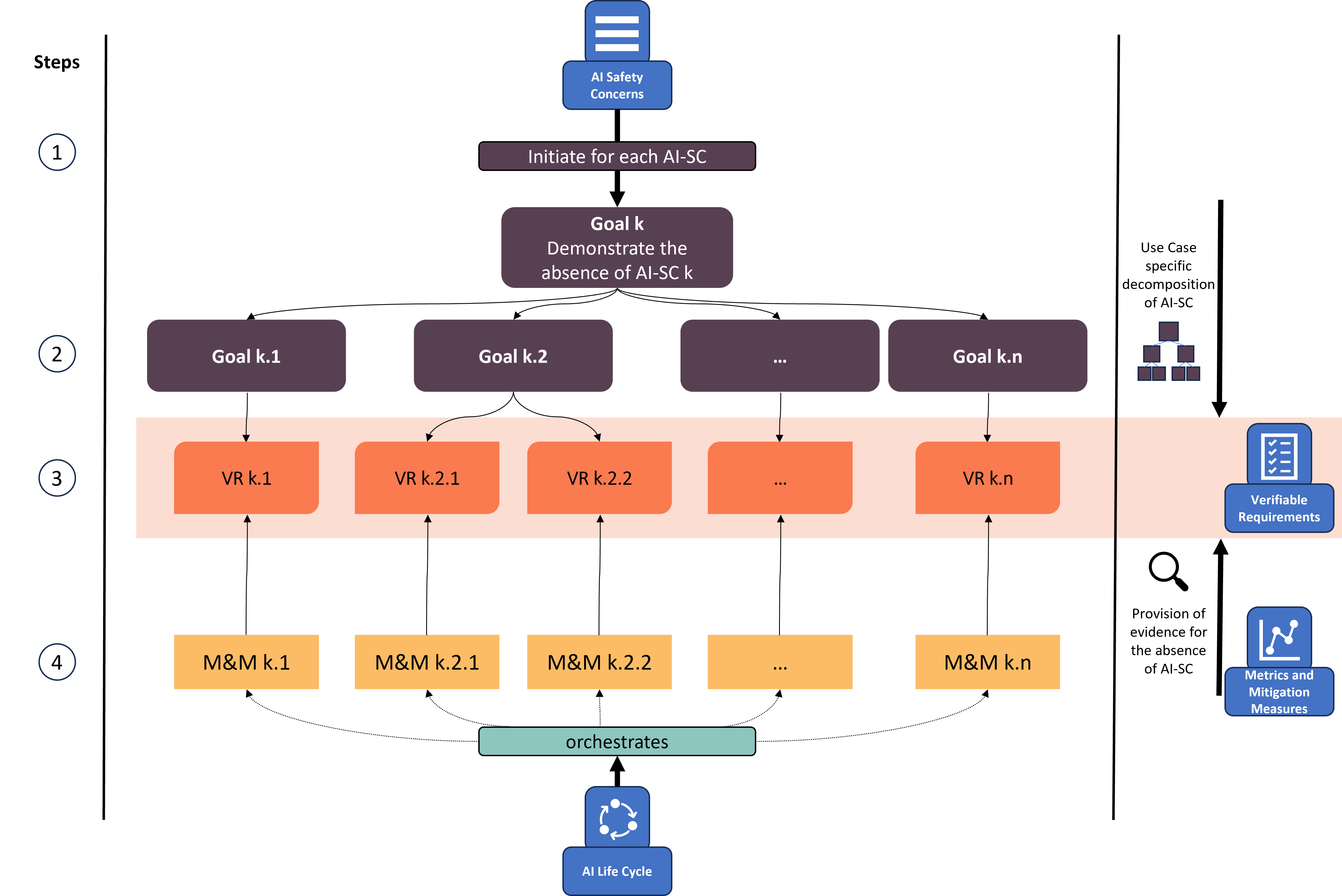}
\caption{LAISC process and argument pattern for demonstrating the absence of AI-SCs, consisting of four steps: 1) Initializing LAISC,2) Decomposing the AI-SC, 3) Derivation of Verifiable Requirements, and 4) Application of Metrics and Mitigation Measures along the  AI life cycle}
    \label{fig: Landscape Logic}
\end{figure*}

This section delineates the detailed instantiation and execution of LAISC methodology, tailored for specific use cases.
A graphical representation of the process, which builds the argument pattern for demonstrating the absence of AI-SCs, is provided in \Cref{fig: Landscape Logic}.

\textbf{1) Initializing LAISC:}
The initiation of the LAISC process involves identifying AI-SCs relevant to the specific use case. 
To start, we recommend building upon the literature to follow the state-of-the-art.
An initial compilation of AI-SCs is presented in \cite{willers_safety_2020}, subsequently expanded in \cite{schnitzer_ai_2024}, which we utilize in the case study in \Cref{sec: Case study}. 
Not every AI-SC may apply universally; for instance, \textit{Problems with synthetic data} is pertinent only if synthetic data is utilized.
This filtering ensures that only relevant AI-SCs are retained, reflecting the specific requirements and safety considerations of the use case. 
Note that the list of AI-SCs is extensible. For instance, new concerns should definitely be added to the list if new concerns are uncovered.
We recommend that to initialize LAISC, a list of AI-SCs relevant to the use case under consideration is developed by an interdisciplinary team of AI and domain experts as well as safety engineers.

\textbf{2) Decomposing the AI-SC:}
The second step in the LAISC process is the decomposition of general AI-SCs into more specific and manageable goals that need to be fulfilled to show the absence of AI-SC.
Since AI-SCs are meant to be applicable across a broad spectrum of AI-based systems, they are initially formulated at a high level of abstraction. 
To argue convincingly for their absence, it is crucial to tailor these AI-SCs to the particular specifications of the use case at hand.

For instance, the AI-SC \textit{Lack of robustness} is universally relevant across AI applications, but the actual requirements for robustness vary significantly depending on the use case specifics.
In the context of a driverless train, weather conditions such as fog or heavy rain significantly impact operational safety.
In contrast, these specific conditions are generally not a concern for autonomous vehicles designed to operate within controlled environments like factory buildings. 
Detailed explanations and examples of how AI-SCs are concretized for the driverless train scenario are provided in \Cref{sec: Case study}.

The decomposition of AI-SCs is continued until a level of detail is achieved, allowing metrics to yield meaningful evidence.
For example, measuring the overall robustness of an AI system is generally impractical due to its broad scope. However, it becomes feasible to assess robustness when it is defined in terms of resistance to specific, well-defined conditions.

\textbf{3) Derivation of Verifiable Requirements:}
Once AI-SCs are decomposed, it is crucial to establish VRs for each. 
These VRs must be explicitly defined to allow for the clear and unambiguous verification of compliance.
For instance, VRs might specify the necessary quality controls in the data labeling process or the thresholds that model outputs must meet under various operational conditions.

Note that, when metrics provide evidence for an AI-SC relating to a model being only a component of the whole system, their impact on overall system safety might not be immediately clear. 
Setting, for instance, definitive thresholds for each quantitative metric could result in inaccurate conclusions and a false sense of confidence in the safety argument.
To mitigate these issues, it is imperative that a team of safety, AI, and domain experts collaboratively evaluate the evidence at a component level, discussing and documenting its adequacy within the assurance case.

\textbf{4) Application of Metrics and Mitigation Measures along the  AI life cycle:}
The final step involves applying appropriate metrics and mitigation measures throughout the AI-based system's life cycle to ensure that all VRs are met.
If initial metrics indicate non-compliance with VRs, mitigation measures are executed to correct deficiencies. 
This cycle of measurement and mitigation continues until compliance is achieved, thereby completing the argument pattern for safety assurance.

\subsection{LAISC Implementation}
\label{subsec: Concrete Application}
A central contribution of LAISC to the assurance of AI-based systems is its ability to interconnect various dimensions of the complex interaction between AI-SCs and their impact on system safety.
For this reason, a tabular representation is an ideal format for LAISC.
In this format, we recommend including columns for each dimension, e.g., AI-SC, AI life cycle stage, VR, M\&M, and system component.
Each row in the table represents a unique combination of an AI-SC and its related dimensions.
For AI-SCs affecting multiple facets of a dimension (e.g., various stages of the AI life cycle or multiple system components), multiple rows are created. 
In systems with substantial complexity featuring numerous AI-based components, the fully instantiated LAISC consists of many rows.
However, applying filters to this table according to specific dimensions facilitates structured analysis. 
For example, suppose a safety engineer needs to focus on issues pertinent to a particular stage of the AI life cycle. In that case, LAISC can be filtered to display only relevant entries.
This comprehensive and organized overview of all necessary actions for assuring specific aspects of the AI-based system increases the methodology's transparency and supports auditing the assurance case.
Note that by providing the tabular structure, we follow the methodology of standards from classical functional safety. For example, IEC 61508 provides recommendations on actions for particular issues in the format of annex tables \cite{IEC_61508}. A snippet of LAISC instantiated for the case study analyzed in \Cref{sec: Case study} in tabular format, is provided in \Cref{tab:LAISC}.
Furthermore, the process of decomposing AI-SCs
can be represented using Goal Structuring Notation (GSN)
[3]. However, since this is not the focus of this paper, we postpone a thorough analysis of the application of GSN to this approach to future work.

\section{Case study: driverless regional train}
\label{sec: Case study}

This section illustrates the application of LAISC to the use case of a driverless regional train.
Concretely, the context depicted in the safe.trAIn research project \cite{zeller_safetrain_2023} is considered. 
The goal of this project is to create a sound safety case for a driverless regional train operating in an open environment.
The whole system developed in the safe.trAIn project is complex and consists of various components, many of them based on AI models.
For simplicity, we consider in the following only a single component, for which the methodology is illustrated:
The track-detector, which is based on a semantic segmentation model that returns a binary segmentation mask for a given camera-based input.
This means that for a given input image, every pixel is assigned the value 1 if it corresponds to the track in the image and 0 if it does not. For this section, we use the AI life cycle model described in \cite{schnitzer_ai_2024}.

Note that due to space restrictions, the following section does not provide a comprehensive application and instantiation of LAISC (covering all AI-SCs).
However, by explicitly describing the methodology along the AI-SCs \textit{Inaccurate data labels}, \textit{Problems with synthetic data}, and \textit{Lack of robustness}, we can showcase the concept since these AI-SCs are considered especially critical for this application. 
For these AI-SCs, we illustrate applying the LAISC process introduced in the previous section: We first explain their relevance for the use case, describe how to decompose the AI-SC under consideration into respective top-level goals, derive verifiable requirements, and finally describe the application of M\&Ms, providing the evidence for the fulfillment of VRs, and ultimately demonstrate the absence of AI-SCs.
The results are summarized in \Cref{tab:LAISC}.
\subsection{Inaccurate Data Labels}
\label{subsec: CS_labels}

\subsubsection{Relevance and Problem Setting}

Supervised learning models, such as the track detector model discussed in this case study, derive their behavior from labeled data.
Specifically, the model utilizes labels in the form of semantic segmentation masks, i.e. for each training image, the label consists of the information for each pixel if it belongs to the railway.
Since these labels represent the ground truth the model aims to learn, the accuracy of these labels is crucial.
Inaccurate labels could prevent the model from learning its intended functionality, making precise data labels indispensable for training a safe AI model.
Existing literature emphasizes the critical importance of avoiding systematic labeling errors, as neural networks have demonstrated the capacity for robust learning despite the presence of random errors \cite{nakkiran_deep_2019,power_grokking_2022,rolnick_deep_2018}. 
Systematic errors pose a significant challenge because they are more likely to induce the model to learn incorrect patterns. 
The exact effect of mislabelled data on the function of a neural network remains an open problem.
The following decomposition aims to handle random labeling errors effectively while eliminating systematic 
errors.

\subsubsection{Decomposing the AI-SC}

To ensure that labeling defects are controlled in such a manner that the model's function doesn't generate unacceptably high errors during operation, three main strategies are employed:
First, it must be ensured the data labeling process meets certain quality standards, i.e., the labeling process follows clearly defined guidelines, which ensure the high accuracy of labels in the dataset.
This includes, when labeling is performed manually, the responsible personnel are trained to adhere to the established guidelines.
In contrast, when labeling is automated, the performance of the algorithms must be rigorously evaluated against the rules set in the guidelines.
Second, the accuracy of the labels used for training and testing the model and its effect on the model performance must be assessed. 
Third, if possible, automatically detecting inaccurate labels should be leveraged to either resolve such instances by a human annotator or excluded them from the data set.

This leads to the following three goals:
\begin{itemize}
    \item \textbf{G1.1} There is a data labeling process in place ensuring accurate data labels.
    \item \textbf{G1.2} The accuracy of the data labels and its impact on model performance is empirically validated.
    \item \textbf{G1.3} Inaccurate labels are automatically detected and reasonably resolved
\end{itemize}
Note that G1.1 refers to a qualitative objective, while G1.2 and G1.3 need to be demonstrated quantitatively, which is reflected in the definition of the VRs in the following.


\subsubsection{Derivation of VRs}
To establish VRs for Goal G1.1, it is essential to define criteria that ensure the data labeling process produces accurate labels.
The specifics of these criteria can vary significantly depending on the use case, including factors such as the data format and the intended purpose of the use case.
A fundamental component of any labeling process is the establishment of a clear and consistent definition of the expected labels. 
For example, in the context of creating binary segmentation masks for rail detection, discrepancies may arise if one annotator interprets the task as labeling only the metallic parts of the rail, while another might include the entire track structure. 
Such ambiguities can significantly reduce model performance, highlighting the need for precise label definitions.

Additionally, the selection of the labeling procedure (manual, semi-automatic, or fully automatic) should align with the project's resources and capacities.
When multiple human annotators are involved, it is crucial to delineate clear roles, such as annotator and independent reviewer, to avoid human bias in the labeling process.
These roles help ensure consistency and accuracy, which are critical for training effective AI models. Therefore, for Goal G1.1 we derive the following VRs:
\begin{itemize}
    \item \textbf{VR1.1.1} The data labeling process adheres to established quality assurance guidelines.
    \item \textbf{VR1.1.2} The labeling personnel is tested upon their knowledge of the labeling guideline.
\end{itemize}

While ideally, each label would be manually verified, practical limitations require a sampling approach. 
Consequently, a significant percentage of labels must be manually checked and corrected by a human inspector to ensure overall accuracy.
Additionally, as discussed above, the impact of random variations in the data labels on model performance is still subject to research.
Therefore, we propose to conduct a sensitivity analysis to better understand this aspect, which could be described in the following:
Consider an AI-model $F$,  and datasets $D_{acc}$ and $D_{inacc}$ consisting of images with  accurate and inaccurate labels (e.g. created by augmenting the labels in $D_{acc}$), respectively.
By doing so, particular types of label inaccuracies can be studied in a controlled manner.
Further consider $F_{D_{acc}}$ and $F_{D_{inacc}}$ to be individual instances of $F$ trained on $D_{acc}$ and $D_{inacc}$, respectively.
The mean performance difference between $F_{D_{acc}}$ and $F_{D_{inacc}}$ evaluated on a testset $D_{test}$ (consisting of images with accurate labels, but being independent of $D_{acc}$) then indicates the effect inaccurate labels can have on model performance.
This leads to the following VRs for G1.2:
\begin{itemize}
    \item \textbf{VR1.2.1} More than x\% of data labels is reviewed by a human inspector, where $x$ is a previously defined number between 0 and 100.
    \item \textbf{VR1.2.2} 
    $|PI_{F_{acc}}(D_{test})-PI_{F_{inacc}}(D_{test})|\leq \epsilon_{1.2.2}$ 
\end{itemize}
Here $PI_F(D)$ denotes the performance indicator of a model $F$ evaluated on the dataset $D$, and  $\epsilon_{1.2.2}$ being a previously defined threshold.
The performance indicator $PI$ should be chosen appropriately for the problem setting and type of AI model under consideration.
Since, in this case study, a segmentation model is considered, a suitable choice for the performance metric is based on IoU \cite{garcia-garcia_survey_2018}, i.e., the performance gap between real and synthetic data is measured by the mean IoU (mIoU) on both data sets. 
Note that a limitation of the evidence described in VR1.2.2 is that it makes a statement about the models $F_{acc}$ and $F_{inacc}$.
To transfer to the actually used model, it is important that their training setup (e.g., architecture, weight initialization, etc.) is the same.

To reach G1.3, we propose to leverage automatic measures to detect inaccurate labels. 
For instance, an established method, called Confident Learning (CL), introduces various techniques to detect inconsistent annotations by estimating the joint distribution of observed and true labels\cite{zhang_characterizing_2020}. This method was also applied to image segmentation tasks \cite{northcutt_confident_2021}.
Of course, such techniques come with limitations (e.g. the performance of CL is affected by the choice of an independent model $\theta$ used to estimate the joint distribution of observed and true labels \cite{zhang_characterizing_2020}). However, they provide additional means of detecting inaccurate data labels if applicable.

Let $CLM$ be a metric based on CL that estimates the label accuracy of a given image, 
then the Verifiable Requirement related to G1.3 is derived as:
\begin{itemize}
  \item \textbf{VR1.3}
    Images $(x,y)$ that fulfill $CLM(x,y)<v_{1.3}$ for a given threshold $v_{1.3}$ and a metric $CLM$, based on CL, are either excluded from the data set or revised by a human annotator.
\end{itemize}

\subsubsection{Application of M\&Ms}
The guidelines for the labeling process strongly depend on the data modality and ML task.
Therefore, to satisfy VR1.1.1, the labeling guidelines should be developed based on existing examples from the literature that are considered to be best practice. 
For instance, the labeling pipeline of the well-known data sets ImageNet \cite{russakovsky_imagenet_2015} or COCO \cite{lin_microsoft_2014} could provide a blueprint. Nevertheless, such procedures should be refined in consultation with domain and data experts to best suit the specific type of data used.
The outcome should include detailed documentation of the labeling process and implemented guidelines.
VR1.1.2 can be fulfilled by evidence showing that the human anotators have the competency to perform the label process in accordance to the guidelines, e.g. by passing a knowledge test.

For VR1.2.1 a report proving that x \% of the data have been revised is sufficient.

Compliance with VR1.2.2 is reached if the metric results fulfill the equation as defined in VR1.2.2.
VR1.3 is fulfilled if all instances that are identified using the metric and threshold defined in VR1.3 are excluded from the dataset or revised by a human annotator.
\subsection{Problems with synthetic data (Reality Gap)}
\label{subsec: CS_Synth}
\subsubsection{Relevance and problem setting}
Modern AI models are trained and evaluated on large datasets. 
Acquiring an adequate number of data points in practical applications, particularly in safety-critical scenarios, is a significant challenge due to the scarcity of real-world data representing scenarios demanding the safety function.
An example of such scenarios are accidents in the railway domain.
Nonetheless, the training and testing of AI-based systems require the use of such data.
One potential solution is the utilization of synthetic data created through simulations, perturbing real data, or employing generative AI models.

However, the use of synthetic data introduces the need to justify their legitimacy, given concerns that discrepancies in image quality between real and synthetic data used for training and testing might mislead the evaluation of the model's operational performance.
This discrepancy, also referred to as \textit{reality gap}, can be formulated as a distribution shift problem with the target domain being the real-world dataset ($D_{real}$) and the initial domain being the synthetically generated dataset ($D_{synth}$) used for training and/or validation of the AI model.
From a safety assurance perspective, it is therefore crucial to demonstrate that the reality gap does not negatively affect the system's safety.
It is important to note that due to the high dimensionality of image data, a straightforward one-to-one or pixel-to-pixel comparison between synthetic and real data points is not feasible.

\subsubsection{Decomposing the AI-SC}
To reach a convincing argumentation for the absence of the reality gap, we distinguish between two aspects:
Firstly, the model is required to reach the same level of performance on real data compared to the synthetic data used for training and testing.
However, secondly, we concluded that evaluating the model's performance on real data is insufficient due to the high dependence used data sets for this comparison.
Additional evidence is required to compensate for the shortcomings of a purely performance-based evaluation.
For that, we aim to demonstrate that the model does not perceive synthetic data systematically different from real data.
This leads us to the following goals:

\begin{itemize}
    \item \textbf{G2.1} The average performance of the model evaluated on real and synthetic data is sufficiently similar.
    
    \item \textbf{G2.2} The model's perception of real and synthetic data is sufficiently similar.
\end{itemize}

\subsubsection{Derivation of VRs}
To derive the VR and related evidence for goal G2.1, it is necessary to specify what "performance" and "sufficiently similar" mean.
With the same argument as for VR1.2.2, the mIoU can be used as a performance metric here.
Since the proposed metric provides numerical values, the question regarding what can be considered "sufficiently similar" is represented by a threshold.
No one-size-fits-all solution exists for determining suitable thresholds on individual metrics. Hence, a diverse team with the required expertise in AI, safety, and domain knowledge derives an appropriate threshold reflecting the state-of-the-art and domain specifics of the use case. 
Consequently, VR2.1 can be formulated:
\begin{itemize}
    \item \textbf{VR2.1} 
    $\mid PI(D_{real}) - PI(D_{synth}) \mid \leq \epsilon_{2.1}$ with $\epsilon_{2.1}$ a prior defined threshold.
\end{itemize}

To derive the VR for G2.2, it needs to be clarified how to measure the perception of the AI model and, again, how to judge if the result of this measurement can be evaluated sufficiently.
Since we can not simply "ask" the AI model how it perceives real and synthetic data, a way to quantify the discrepancy in the model's perception is required.
A methodology from the field of Explainable AI and Out-of-Distribution Detection, called \highlight{Neural Activation Patterns}{NAPs}, aims to derive insights into models' internal mechanisms by analyzing the values of the activation functions during model inference.
Applied to this use case, the activations of neurons at given layers of the network during inference on the synthetic and real datasets are calculated.
Thus, we can compare the two respective distributions.

Suppose the distributions differ (metrics for measuring similarity or the distance between distributions in the context of NAPs are given in \cite{cheng_runtime_2019,cheng_safety_2023}) from each other more than a given threshold. In that case, the synthetic data are deemed questionable due to the reality gap.
The process for deriving the threshold is analogous to VR2.1.

So let $Pr_{NAP}^{real}, Pr_{NAP}^{synth}$ be the distributions of the NAPs for real and synthetic data, respectively, and $\Delta$ be a metric to compare two probability distributions, then VR1.2 can be formulated as:
\begin{itemize}
    \item \textbf{VR2.2} $\Delta(Pr_{NAP}^{real},Pr_{NAP}^{synth}) \leq \epsilon_{2.2}$, with $\epsilon_{2.2}$ a prior defined threshold
\end{itemize}

\subsubsection{Application of M\&Ms}
For VR2.1 and VR2.2 both quantitative VRs were derived, i.e. metrics and corresponding thresholds are specified.
Therefore, applying the metrics and comparing the results to the tresholds allows a clear evaluation whether the VRs are fulfilled.
\begin{table*}[h!]
    \centering
    \begin{tabular}{|p{80pt}|p{100pt}|p{130pt}|p{25pt}|p{90pt}|}
    \hline
       AI-SC&
        Stage in AI Life Cycle&
       Decomposition&
       VR&
       M\&M\\
    \hline
    \hline
        Inaccurate data labels &
        Data collection \& preparation &
        Quality assured labeling process (G1.1)&
        VR1.1.1 &
        Labeling guidelines\\
    \hline
        Inaccurate data labels &
        Data collection \& preparation &
        Quality assured labeling process (G1.1)&
        VR1.1.2 &
        Annotator knowledge test\\
   \hline
         Inaccurate data labels &
         Data collection \& preparation &
         Empirical label validation and effect analysis (G1.2)&
         VR1.2.1 &
         Sampling-based manual label inspection\\
    \hline
         Inaccurate data labels &
         Data collection \& preparation &
         Empirical label validation and effect analysis (G1.2)&
         VR1.2.2 &
         Label sensitivity analysis\\
    \hline
         Inaccurate data labels &
         Data collection \& preparation &
        Automatic detection of inaccurate labels (G1.3)&
         VR1.3 &
         Confident Learning metric\\
    \hline
        Problems with synthetic data &
        Data collection \& preparation &
        Reality gap in the model's performance (G2.1)&
        VR2.1 &
        Reality gap performance test\\
        \hline
        Problems with synthetic data &
        Data collection \& preparation &
        Reality gap in the model's perception (G2.2)&
        VR2.2&
        Neural activation metric \\
    \hline
        Lack of robustness &
        Modeling &
        Lack of adversarial robustness (G3.1)&
        VR3.1&
        Argumentation\\
        \hline
        Lack of robustness &
        Modeling &
        Lack of robustness against natural variation in the input domain (G3.2)&
        VR3.2&
        Perturbation-based robustness quantification\\
    \hline
        Lack of robustness&
        Modeling&
        Lack of robustness against counterfactuals (G3.3)&
        VR3.3&
        Counterfactual analysis\\
    \hline
    \end{tabular}
    \vspace{1mm}
          \caption{Excerpt of laisc filtered for the AI safety concerns discussed in \Cref{sec: Case study}.}  
    \label{tab:LAISC}
\end{table*}
\subsection{Lack of robustness}
\label{subsec: CS_robustness}
\subsubsection{Relevance and problem setting}
According to ISO IEC 22989, robustness for AI systems describes "the ability to maintain their level of performance, as intended by the developers, under any circumstances" \cite{ISO_22989}.
The apparent necessity for robust AI systems is also exemplified through the proposed regulation for AI systems (European AI Act \cite{AI_Act}), which explicitly requires high-risk AI systems to "achieve, in the light of their intended purpose, an appropriate level of [...] robustness".
Since the AI-based system should be able to operate robustly even when confronted with disturbances in the input domain, such as dust on the lens or adversarial weather conditions, in this use case, the demonstration of the absence of this AI-SC is crucial for a sound safety case.

\subsubsection{Decomposing the AI-SC}
To demonstrate the absence of \textit{Lack of robustness}, we need to explicitly define the circumstances requiring system robustness and what level of robustness is considered sufficient.
To do so, we break down the AI-SC \textit{Lack of robustness} into two main aspects:
Firstly, the robustness against malicious attacks to the system, called \textit{adversarial attacks}, and secondly, naturally occurring variations in the input domain caused by changes in the operating environment or system itself, such as sensor noise.

Regarding the first aspect, adversarial attacks include maliciously created pixel-wise perturbations of input images as well as the inclusion of motives in input images that can affect the model's performance.
The consideration of such cases is excluded from the use case with the following argumentation:

The system's risk analysis evaluated the likelihood of an attacker specifically creating adversarial examples to hijack the system's behavior as negligibly small.
Furthermore, from a technical standpoint, the benefit of such testing and robustification of a model is still questionable even among academia \cite{carlini_towards_2017}.
Therefore, we conclude that the required evidence for the absence of \textit{Lack of robustness} is restricted to the aspects of naturally occurring input variations.

To study the robustness of a given AI component under natural perturbations, proper modeling of the operational environment is required.
For this use case, the environment is modeled by specifying the Operational Design Domain (ODD), and we also assume the system specifications to be given.
Note that the inadequate specification of the ODD can be considered an AI-SC \cite{schnitzer_ai_2024}. For details on demonstrating the absence of \textit{Inadequate specification of ODD} we refer to \cite{weiss_approach_2024}.
Consequently, for the following discussion, we assume a properly defined ODD. Based on the information therein, a list of operational conditions can be derived against which the system shall be robust.
In this use case, the relevant conditions are sensor noise, adverse lighting conditions, external sensor factors (e.g., dust, droplets, lens defects, etc.), adverse weather conditions, partial occlusions, and geometrical transformations.

A limitation of this approach is the difficulty in asserting the completeness of the identified conditions.
Thus, it is crucial to acknowledge that beyond testing robustness against specifically defined conditions, developing a strategy for identifying new, unknown conditions that could lead to performance degradation is additionally required.
We call the analysis aiming to identify such conditions counterfactual analysis.
To summarize, the concretization of the AI-SC \textit{Lack of robustness} yielded the following sub-goals:
\begin{itemize}
    \item \textbf{G3.1} It can be argued that robustness against adversarial attacks is considered not relevant for the use case.
    \item \textbf{G3.2} The track detector is sufficiently robust against adversarial conditions specified in the ODD.
    \item \textbf{G3.3} A counterfactual analysis has shown no unknown weaknesses in terms of the model's robustness.
\end{itemize}

\subsubsection{Derivation of VRs}
Since G3.1 requires an argumentation, the related VR can be simply formulated as:
\begin{itemize}
    \item \textbf{VR3.1} An argumentation why the absence of adversarial robustness is not required can be provided and is approved by a team holding expertise on safety, AI, and domain knowledge.
\end{itemize}
To provide the necessary evidence for G3.2, a systematic approach for quantifying the performance under specific conditions, such as adverse lighting, is essential.
Given the scarcity of real data for all relevant conditions, this data is simulated by introducing perturbations to real data instances.
For example, altering an image's brightness and contrast can simulate various lighting scenarios.
However, as discussed in \Cref{subsec: CS_Synth}, the ability to reflect real data needs to be justified.
Once a dataset $D_i$ representing a particular condition is created, its impact on model performance can be evaluated using an appropriate performance indicator $PI$, such as mIoU for the track detector.
To conclude the derivation of VR3.2, it is crucial to define what constitutes sufficient robustness.
Given that the metric provides a numerical value, a suitable threshold must be determined, taking into account the likelihood and severity of the respective condition.

Similarly, to derive VR3.3, a dataset containing counterfactual examples needs to be created and a suitable threshold determined.
\begin{itemize}
    \item \textbf{VR3.2} $PI(D_i) \geq v_{3.2}^i \ \forall i$. where each $i$ represents a condition defined in the ODD and $v_{3.2}^i$ respective prior determined threshold.
    \item \textbf{VR3.3} $PI(D_{cf}) \geq v_{3.3}$ where $D_{cf}$ is the data set created during counter factual analysis and $v_{3.3}$ is a prior determined thresholds. 
\end{itemize}
\subsubsection{ Application of M\&Ms}
Since VR3.1 explicitly requires an argumentation, a document providing this argumentation forms the required evidence.
VR3.2 and VR3.3 are again constructed by applying a metric to specific data sets; compliance with these VRs can be checked by comparing metric results 
to the related thresholds.

\subsection{Remarks}
Whenever VRs and related M\&Ms analyze differences in performance caused by particular attributes (such as the label accuracy in VR1.2.2 or the reality gap in VR2.1), it is crucial that the respective datasets are similar in terms of other semantic attributes, i.e., the proportion of images with certain weather conditions, objects on the rail, etc. 
Also, the datasets should include enough data points to provide statistically significant results. 
Finally, note that we do not claim that evidence from the specific metrics described above is sufficient to demonstrate the absence of the discussed AI-SCs. 
This section aimed to illustrate the methodology of how such evidence can be derived to support the assurance case using LAISC.


\section{Discussion and Limitations}
\label{sec: Discussion}

In \Cref{sec: Landscape}, we introduced the Landscape of AI Safety Concerns and demonstrated its utility in supporting the development of safety assurance cases for AI-based systems.
A primary advantage of this approach is its systematic nature, deriving evidence for demonstrating the absence of AI safety concerns.

Since autonomous systems are the most relevant class of systems in safety-critical applications, this was the focus of our work.
However, we believe that the concept of LAISC also generalizes to safety-relevant, AI-based systems in general.


Furthermore, while our methodology significantly advances the structured development of assurance cases for AI-based systems, it does not provide a universal guarantee of safety. 
LAISC effectively demonstrates how to structure evidence for the absence of AI-SCs, but addressing these concerns alone  does not provide estimations for the system’s probability of causing harm.
Rather, the application of LAISC primarily strengthens confidence in the safety assurance process by demonstrating that AI-specific issues are managed effectively.

Consequently, we view LAISC as one of several critical components contributing to an AI-based system's comprehensive assurance case. 
This holistic approach acknowledges that while LAISC strengthens the safety assurance case by addressing specific AI-related issues, it must be integrated with other safety assurance measures to ensure a comprehensive evaluation of overall system safety.


Lastly, decomposing abstract AI-SCs into measurable goals is vital for linking concrete evidence with overarching goals in our methodology.
Yet, this decomposition introduces the risk of information loss, i.e., it needs to be assured that the process of decomposition is complete and sound.
However, note that the challenge of ensuring the conjunction of sub-goals fully captures the essence of the higher-level goals is typical in top-down structured assurance cases. 
Additionally, providing concrete guidance for breaking down top-level goals into finer details is impractical in a general setting due to the dependency on specific use case conditions.
We intentionally refrained from defining these details to maintain the methodology's flexibility and applicability across various applications.

\section{Conclusion and future work}
\label{sec: Conclusion}

In this paper, we tackled the challenges associated with assuring the safety of AI-based systems.
To address these challenges, we introduced a novel methodology that supports the development of safety assurance cases for AI-based systems based on evidence demonstrating the absence of AI safety concerns.
This evidence is produced by the derivation of application-specific and Verifiable Requirements and the application of Metrics and Mitigation Measures showing the achievement of the VRs.
We demonstrated the practical application of this approach with a safety-critical use case of a driverless regional train.

For future research, we plan to apply the methodology to more use cases, enhance the methodology, and further evaluate its effectiveness.
For instance, this will provide more insights into determining the thresholds for numerical M\&Ms.
This area presents a significant opportunity for further development and refinement in the field of AI safety assurance.

\section*{Acknowledgement}
This research has received funding from the Federal Ministry for Economic Affairs and Climate Action (BMWK) under grant agreements 19I21039A.
We also want to thank Marc Zeller and Andreas Hapfelmeier for their valuable discussions and feedback, which contributed to improving this research.
This version of the manuscript was submitted by the 8th International Conference on System Safety and Reliability (ICSRS). The authors applied minor adaptions to this version based on the reviewers' feedback, which will be reflected in the conference proceedings.
\bibliographystyle{splncs04}
\bibliography{references.bib}

\end{document}